\documentclass[conference]{IEEEtran}
\IEEEoverridecommandlockouts

\usepackage{cite}
\usepackage{array}
\usepackage{amsmath,amssymb,amsfonts}
\usepackage{graphicx}
\usepackage{textcomp}
\usepackage{xcolor}
\usepackage{makecell}
\usepackage{url}
\usepackage{hyperref}
\usepackage{multirow}
\usepackage{algorithm}
\usepackage{algpseudocode}
\usepackage{tikz}

\def\BibTeX{{\rm B\kern-.05em{\sc i\kern-.025em b}\kern-.08em
    T\kern-.1667em\lower.7ex\hbox{E}\kern-.125emX}}

\newcommand{\copyrighttext}{%
  \footnotesize
  \textcopyright\ 2025 IEEE. Personal use of this material is permitted. Permission from IEEE must be obtained for all other uses, in any current or future media, including reprinting/republishing this material for advertising or promotional purposes, creating new collective works, for resale or redistribution to servers or lists, or reuse of any copyrighted component of this work in other works.}

\newcommand{\copyrightnotice}{%
  \begin{tikzpicture}[remember picture,overlay]
    \node[anchor=south,yshift=12pt] at (current page.south)
      {\fbox{\parbox{\dimexpr0.72\textwidth-\fboxsep-\fboxrule\relax}{\copyrighttext}}};
  \end{tikzpicture}%
}

\begin{document}
\pagestyle{plain}
\pagenumbering{arabic} 
\setcounter{page}{1} 


\makeatletter
\newcommand{\linebreakand}{%
    \end{@IEEEauthorhalign}
    \hfill\mbox{}\par
    \mbox{}\hfill\begin{@IEEEauthorhalign}
}
\makeatother

\title{Evolutionary Hyperparameter Optimization to Find Lightweight CNN Models for Autonomous Steering}

\author{
    \IEEEauthorblockN{Devson Butani}
    \IEEEauthorblockA{\textit{Dept. of Math and Computer Science} \\
        \textit{Lawrence Technological University}\\
        Southfield, MI, USA \\
        dbutani@ltu.edu}
    \and
    \IEEEauthorblockN{Ryan Kaddis}
    \IEEEauthorblockA{\textit{Dept. of Math and Computer Science} \\
        \textit{Lawrence Technological University}\\
        Southfield, MI, USA \\
        rkaddis@ltu.edu}
    \and
    \IEEEauthorblockN{Chan-Jin Chung}
    \IEEEauthorblockA{\textit{Dept. of Math and Computer Science} \\
        \textit{Lawrence Technological University}\\
        Southfield, MI, USA \\
        cchung@ltu.edu}
}

\maketitle

\noindent{\small This work has been accepted for publication in \emph{2025 IEEE International Conference on Electro Information Technology (eIT)}. 
This is the author-accepted manuscript version. 
The final published version is available at \url{https://doi.org/10.1109/eIT64391.2025.11103679}.}
\vspace{0.8em}
\copyrightnotice

\begin{abstract}
    This research investigates the optimization of Convolutional and Dense Neural Networks (CNNs and DNNs) for autonomous steering using the (N+M) Evolution Strategy (ES) with the 1/5th success rule. The primary objective is to develop a lightweight CNN based model capable of real-time steering angle prediction, mimicking human driving behavior on predefined paths. The ES algorithm automates hyperparameter tuning, dynamically adjusting parameters such as filter sizes and layer configurations. Data collection encompasses driving scenarios recorded via the LTU ACTor autonomous driving platform, including variations in path direction and driving style. The very small dataset consists of timestamped images labeled with steering angles and pre-processed to focus on relevant visual information. Initial experiments involve training a baseline CNN model, which is then refined using ES to significantly reduce the size of the model while maintaining competitive predictive accuracy. The results highlight the viability of lightweight neural network architectures for real-time autonomous systems, striking a balance between computational efficiency and performance. This study not only advances research initiatives on the use of evolutionary algorithms for autonomous driving applications but also lays the foundation for the deployment of cost-effective and scalable solutions in self-driving technology.
\end{abstract}


\section{Introduction}
Autonomous driving systems require robust models capable of making real-time decisions under varying conditions. A critical component of these systems is the prediction of steering angles based on camera input, which involves processing visual data effectively and efficiently on compute-limited onboard processors~\cite{PilotNET_application}. This research focuses on training and optimizing CNNs using Evolutionary Strategy (ES) to automate the search for the best performing and best size-reduced model configurations.

This research directly addresses the limitations that come with larger models and limited data. Larger models, while often achieving higher accuracy, come with increased computational cost and memory requirements, which can lead to slower inference times and higher energy consumption~\cite{self_drive_latency_and_model_size_drawbacks}. This poses a significant challenge for real-time applications like autonomous driving, where split-second decisions and energy efficiency are crucial. The size of a model directly impacts its feasibility for deployment, particularly in resource-constrained onboard processors~\cite{MobileNET_size_and_speed}. Minimizing the size of the model and also training with a very small dataset is the core design challenge in autonomous driving systems.

Furthermore, the performance of deep learning models is highly dependent on the availability of large, diverse, and labeled datasets. In the context of autonomous driving, collecting and annotating such datasets for every possible scenario is impractical and cost prohibitive~\cite{ImageNET_large_scale_database}. Although small datasets can lead to over-fitting, where the model performs well on the training data but fails to generalize to unseen scenarios or conditions, this is often not a problem in real-world applications where the operational design domain (i.e., the area of interest) can be regulated. For example, deploying an autonomous vehicle specifically for a small city or a specific set of routes. With this method, we can leverage the addition of small data sets every time the domain expands and utilize techniques like transfer learning and few shot learning to improve model performance over time~\cite{few_shot_learning}. Although optimization techniques have been extensively explored in various domains, their application to creating lightweight, well-generalized, and adaptable models for autonomous driving, specifically for real-time steering angle prediction, remains an active area of research.

Evolution Strategy (ES), one of the Evolutionary Algorithms (EAs)~\cite{ES_introduction}, offers a compelling advantage in this context. ES is a population-based optimization algorithm that leverages the collective intelligence of a population of candidate solutions to find the best-performing solution within a given search space~\cite{ES_introduction}. Unlike gradient-based optimization methods, which can struggle in complex, non-differentiable, or noisy search spaces, ES algorithms are well-suited for exploring such landscapes~\cite{ES_scalable}. 

In contrast to knowledge distillation techniques~\cite{distillation} that train smaller models to mimic larger pre-trained ones, our evolutionary approach directly finds the architectural hyperparameters to discover inherently efficient and small models within the search space. Compared to distillation, ES eliminates the need for large datasets and to design the two working model architectures to transfer knowledge between. This is particularly relevant to hyperparameter optimization of CNNs for autonomous driving, where the relationship between operational domain, model architecture, hyperparameters, and performance can be highly complex and non-linear.

\subsection{Data Acquisition and Pre-processing}

\textbf{Equipment:} The LTU ACTor autonomous driving platform, integrated with the Robot Operating System (ROS), was used for data collection. Sensor data, including images from the vehicle's forward-facing camera and corresponding steering angles, were recorded in rosbag files. This setup ensured synchronized visual and control data, essential for training and evaluating the models.

\textbf{Environment:} Data was collected along a predefined circular path: the red brick path surrounding Ockham's Wedge at LTU. To introduce variability and enhance model robustness, driving sessions included clockwise and counterclockwise directions, smooth and zigzag maneuvers, and driving along inner and outer path edges to simulate diverse spatial alignments. Figure~\ref{fig:ockham} shows an overhead view of the data collection site.

\begin{figure}[ht]
    \centering
    \includegraphics[width=2in]{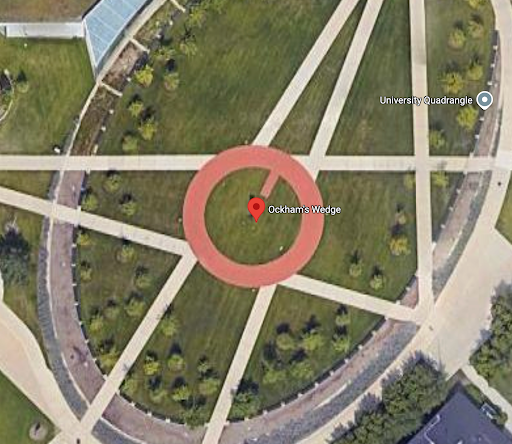}
    \caption{Google Maps aerial view of Ockham's Wedge at Lawrence Technological University. Data was collected from the red brick circle surrounding the art piece.}
    \label{fig:ockham}
\end{figure}

\textbf{Extraction and Pre-processing:} A custom script was developed to extract and pre-process data from ROS rosbag files.

\begin{figure}[ht]
    \centering
    \includegraphics[width=2.0in]{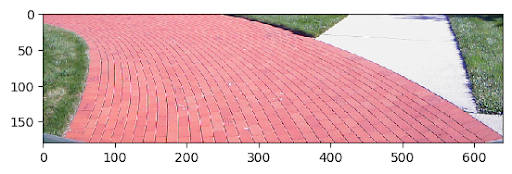}
    \includegraphics[width=2.0in]{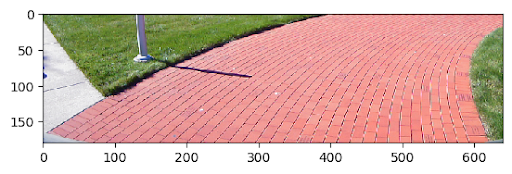}
    \caption{Sample images after cropping and pre-processing. The extracted steering angles are 12.901 degrees (Top) and -10.099 degrees (Bottom).}
    \label{fig:preprocessed_images}
\end{figure}

Images were saved at 200 ms intervals, with filenames encoding timestamps and steering angles. Each image was cropped to exclude irrelevant regions, such as the sky and surrounding buildings. The resulting dataset consisted of only 2,957 images, split into 70\% training (2,069), 20\% validation (592), and 10\% testing (296). Augmented images were not added to this small dataset. The smaller dataset enables this proof-of-concept experiment on available hardware; it is well known that expanding the dataset—especially to include varied lighting and weather—would slow down training and optimization while improving model generalization and performance. Even though the dataset is small, it is representative of real-world driving scenarios and provides a useful benchmark for evaluating model performance.

\subsection{Research Goals}

Since there are many experiments in image to steering angle prediction that have shown quality improvements over the years~\cite{PilotNET_application, lane_detection_good_results_zigzag, CNN_can_self_drive, river23book, CNN_RNN_Steer}, the proposed ES based optimization should allow for a significant improvement in deployable performance and ability to run real-time inference on on-board compute resources. The primary goals of this research are to:

\begin{enumerate}
    \item Develop a framework to apply the (N+M) Evolution Strategy (ES) with the 1/5th success rule for automated hyperparameter tuning.
    \item Minimize the size of the baseline CNN model while maintaining satisfactory steering angle prediction performance.
    \item Validate the real-world applicability of ES-optimized models in autonomous vehicle systems.
\end{enumerate}

\section{Methodology}

\subsection{Model Training and Baseline Establishment}

Models are trained using the Keras API with a PyTorch backend to leverage GPU compute capacity. The training process employs the Mean Squared Error (MSE) loss function, which quantifies the average squared difference between the predicted and actual steering angles. The Mean Absolute Error (MAE) serves as a key performance metric, indicating the average deviation in steering angle predictions in degrees. This provides a straightforward and interpretable measure of model accuracy for autonomous steering applications. Minimizing both MSE and MAE is crucial to achieving precise control of the vehicle. These metrics serve as benchmarks for evaluating the efficacy of ES with the 1/5 success rule in optimizing model architectures.

Initial experiments began with a single-layer CNN to establish a rudimentary baseline. However, the limited capacity of this architecture renders it unsuitable for real-world driving scenarios. To address this, the PilotNET architecture, a CNN and DNN combination developed by NVIDIA~\cite{PilotNET} was adopted. 

\begin{figure}[ht]
    \centering
    \includegraphics[width=3in]{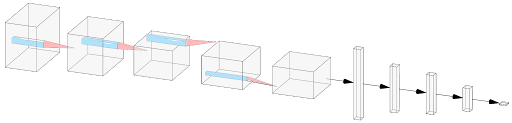}
    \caption{Visual representation of the baseline model architecture using~\cite{LeNail2019}}
    \label{fig:alexnet}
\end{figure}

PilotNET, an early milestone in autonomous steering research, demonstrated the potential of GPU-accelerated deep learning for this domain. This simple architecture, shown in Figure~\ref{fig:alexnet}, works as the baseline because this study is benchmarking performance improvements rather than absolute performance. Leveraging this proven architecture allowed for a more effective comparison of ES-optimized models against a recognized standard.

Early stopping was implemented to improve training efficiency and prevent overfitting. Training instances are terminated if no improvement in the MSE metric is observed after four epochs. This strategy prevents the allocation of computational resources to hyperparameters that do not contribute to model performance, thereby accelerating the optimization process. 

\subsection{Hyperparameter Management}

Initially, the hyperparameter space included individual layer units (number of filters in convolutional layers and neurons in fully connected layers), batch sizes, learning rates, activation functions, and optimizer selection. To facilitate a comprehensive exploration of potential architectures, the layer units were allowed to vary from 20\% to 300\% of the PilotNET baseline.

\begin{table}[ht]
    \centering
    \caption{Hyperparameter Search Spaces}
    \begin{tabular}{r|l}
        \textbf{Hyperparameter} & \textbf{Search Space}     \\
        \hline
        CNN filter and DNN neuron units  & 20\% to 300\% of PilotNET \\
        Batch Size              & 1 - 32                    \\
        Learning Rate           & 0.001 - 1.5               \\
        Activation Functions     & ReLU, eLU, Sigmoid, Tanh  \\
        Optimizers               & Adam, SGD, RMSprop        \\
    \end{tabular}
    \label{tab:searchspace}
\end{table}

However, preliminary experiments indicated that focusing optimization efforts solely on layer units resulted in improved performance and reduced search time. Consequently, batch size (32), learning rate (0.001), activation function (ReLU), and optimizer (Adam) were fixed to match the baseline model configuration. This narrowed scope allowed ES to concentrate on the most impactful architectural parameters, leading to more efficient exploration of the design space. The ranges of the layer units explored are provided in Table~\ref{tab:searchspace}.

\subsection{Optimization Framework}

CNN hyperparameters, specifically the number of filters in convolutional layers and the number of neurons in fully connected layers, are encoded as real-valued genes within the ES framework~\cite{es_github}. The (N+M)-ES approach iteratively optimizes hyperparameters by combining the N best-performing parent models with M offspring models generated through mutation. This approach allows for both exploitation of promising regions of the hyperparameter space (through selection of top parents) and exploration of new possibilities (through mutation of offspring).

\subsection{ES Algorithm with 1/5 Success Rule}

The Evolution Strategy (ES) algorithm employs the 1/5th success rule~\cite{ES_one_fifth_rule} to dynamically adjust the mutation step size (\(\sigma\)). This adaptive mechanism ensures efficient exploration of the hyperparameter space while avoiding premature convergence to local optima. The rule is defined as follows:
\begin{equation}
  \sigma_{t+1} = \left\{
  \begin{array}{ll}
      \alpha * \sigma_t & \text{success rate} > \frac{1}{5}             \\
      \beta * \sigma_t  & \text{success rate} \leq \frac{1}{5}          \\
  \end{array}
  \right.
\end{equation}
where \(\sigma_{t+1}\) is the updated step size for the next generation. \(\sigma_t\) is the current step size. \(\alpha > 1\) is a scaling factor to increase the step size (e.g., 1.22). \(\beta < 1\) is a scaling factor to decrease the step size (e.g., 0.82). The "success rate" is defined as the fraction of offspring that outperform their parents in terms of validation performance.

If the optimization success rate is greater than 1/5th within a generational window, the step size is increased, encouraging greater exploration of the hyperparameter space. Conversely, if the success rate is less than or equal to 1/5, the step size is decreased, promoting more focused refinement of promising solutions. This adaptive mechanism ensures efficient exploration while avoiding premature convergence to local optima.

The ES optimization process with 1/5 success rule~\cite{chung1997knowledge, chung1998caep, chung2000knowledge} is as follows:

\begin{enumerate}
    \item \textbf{Initialization:} Generate, train and evaluate N parent models with random hyperparameters sampled within predefined ranges. The initial hyperparameters are drawn from a uniform distribution within the specified ranges. The number of parents, N, is a tunable parameter that controls the diversity of the population.
    
    \item \textbf{Child Generation:} Create M offspring by mutating parent hyperparameters according to the following equation:
          \begin{equation}
              \hat{h} = h + \text{gauss}(0, \sigma \cdot (max(h) - min(h)))
          \end{equation}
          where \(\hat{h}\) represents the mutated hyperparameter value. \(h\) is the original hyperparameter value from the parent model. \(\text{gauss}(\mu, \sigma)\) is a random number drawn from a Gaussian distribution with mean \(\mu\) and standard deviation \(\sigma\).  In this case, \(\mu = 0\). \(\sigma\) is the mutation step size, controlling the magnitude of the mutation. \(\text{max}(h)\) and \(\text{min}(h)\) define the upper and lower bounds of the hyperparameter search space, respectively.

          ~~The term \((\text{max}(h) - \text{min}(h))\) normalizes the mutation step size to the range of the hyperparameter, ensuring that the mutation is proportional to the scale of the parameter search space.
          
    \item \textbf{Training and Evaluation:} Train all offspring models and evaluate their performance based on validation loss (MSE) and accuracy (MAE). These are calculated using the test split of the dataset.
    
    \item \textbf{Selection and Step size Adjustment:} Retain the top N models (from parents and children) based on validation performance to form the next generation's parent population. A window size of 5 generations was used to balance exploitation and exploration for this particular experiment as seen in Algorithm~\ref{alg:es}. This selection process ensures that the most promising models are carried forward, driving the evolution towards improved performance while also allowing exploration to nearby search space. The mutation step size (\(\sigma\)) is dynamically adjusted using the 1/5 success rule~\cite{ES_one_fifth_rule}. 
\end{enumerate}

As described in Algorithm 1, this process repeats until the max generation count is reached or the termination condition, a solution much better than the baseline, is met; for this research max generations were limited to 100 and the termination condition was a MAE of 0.1 degrees.

\begin{algorithm}
\caption{(N+M) Evolution Strategy with 1/5 Success Rule}
\label{alg:es}
\textbf{Input:} Hyperparameter ranges $[min(h), max(h)]$, number of parents $N$, offspring $M$, generations $G$, step size factors $\alpha > 1$, $\beta < 1$, window size $W$ \\
\textbf{Output:} Best model with optimized hyperparameters
\begin{algorithmic}
\State Initialize, train and evaluate $N$ parents random parent models
\State Set step size $\sigma \gets$ initial value, generation $t \gets 0$, positive counter $P \gets 0$
\While{$t < G$ and termination condition not met}
    \State Generate $M$ offspring:
    \[
    \hat{h} = h + \text{gauss}(0, \sigma \cdot (max(h) - min(h)))
    \]
    \State Train and evaluate offspring
    \State Combine parents and offspring
    \State Select top $N$ models as next-generation parents
    \If{at least one offspring outperforms parents}
        \State Increment $P \gets P + 1$
    \EndIf
    \If{$t$ mod $W == 0$} \Comment{Adjust $\sigma$ every $W$ gen.}
        \State Calculate success rate:
        \[
        \text{success rate} = P / M
        \]
        \If{$\text{success rate} > 1/5$}
            $\sigma = \alpha \cdot \sigma$
        \Else
            ~$\sigma = \beta \cdot \sigma$
        \EndIf
        \State Reset $P \gets 0$
    \EndIf
    \State Increment $t$
\EndWhile
\State Return best model from final population
\end{algorithmic}
\end{algorithm}

\textbf{Computational Resources: } Training is performed on Lawrence Technological University's NVIDIA A100 GPU server, enabling efficient parallel training of N parent and M offspring model pools. The large VRAM (80GB) capacity of the A100 GPU allows for the use of larger N and M values, which increases the diversity of the population and enhances the exploration of the hyperparameter space. The specific values of N and M, shown in Table~\ref{tab:testarch} are based on the availability of shared computational resources during training and the complexity of the model. While other N and M values (1 - 10) were used, Table~\ref{tab:testarch} values are from the optimization runs that showed the best performance. One of the benefits of reducing model size is that it allows ES optimization with higher N and M values or free up resources for other concurrent projects.

\subsection{Model Selection}

Four different model architectures are proposed to determine the effect of ES on model performance and size. The first model, named "Baseline", is the same architecture established by PilotNET~\cite{PilotNET}. The second model, named "Optimized", contains a CNN and DNN structure that has been optimized with ES, within the search space proposed in Table~\ref{tab:searchspace}. The third model, named "Half-Size", contains the PilotNET CNN architecture with half of the baseline number of CNN filters per layer, and half of the DNN search space proposed by Table~\ref{tab:searchspace}. The fourth model, named "Quarter-Size", contains the PilotNET CNN architecture with a quarter of the baseline number of CNN filters per layer, and a quarter of the DNN search space proposed by Table~\ref{tab:searchspace}. Note that the pretrained weights of PilotNET were not used in this study. These four models allow the effect of ES on model performance with smaller model sizes to be observed.

\begin{table}[ht]
    \centering
    \caption{Model Descriptions}
    \begin{tabular}{p{0.75in}|p{2.25in}}
        \textbf{Model Name} & \textbf{Description}                                                 \\
        \hline
        Baseline            & PilotNET architecture, baseline for comparison.                      \\
        Optimized           & ES-optimized CNN \& DNN layers.                                      \\
        Half-Size           & Half the size of baseline CNN layers and ES-optimized DNN layers.    \\
        Quarter-Size        & Quarter the size of baseline CNN layers and ES-optimized DNN layers. \\
    \end{tabular}
    \label{tab:aliases}
\end{table}

\subsection{Model Testing}

To validate the real-time applicability of the ES-optimized models, a series of tests were conducted within a 2D simulation environment, GazelleSim~\cite{gazelle_sim}. The simulation allows for controlled and repeatable testing of the steering control algorithms on the red brick circular path (operational design domain) used for training. For variance, both directions (clockwise and counterclockwise) were tested; however, the difference in performance was negligible.

\begin{figure}[ht]
    \centering
    \includegraphics[width=2.5in]{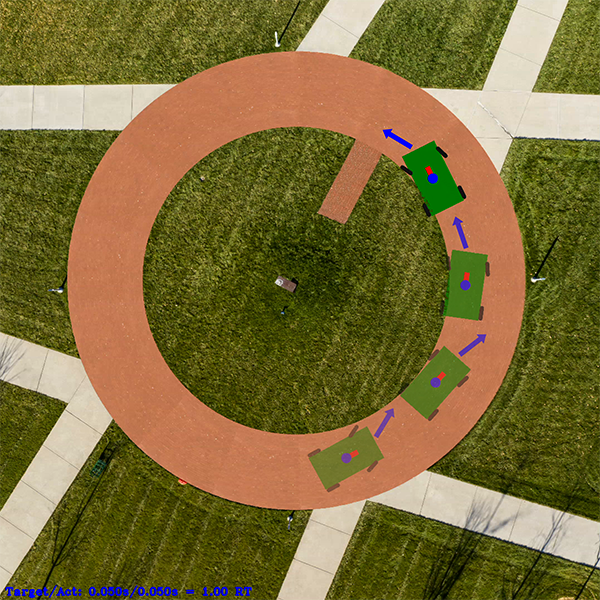}
    \includegraphics[width=3.49in]{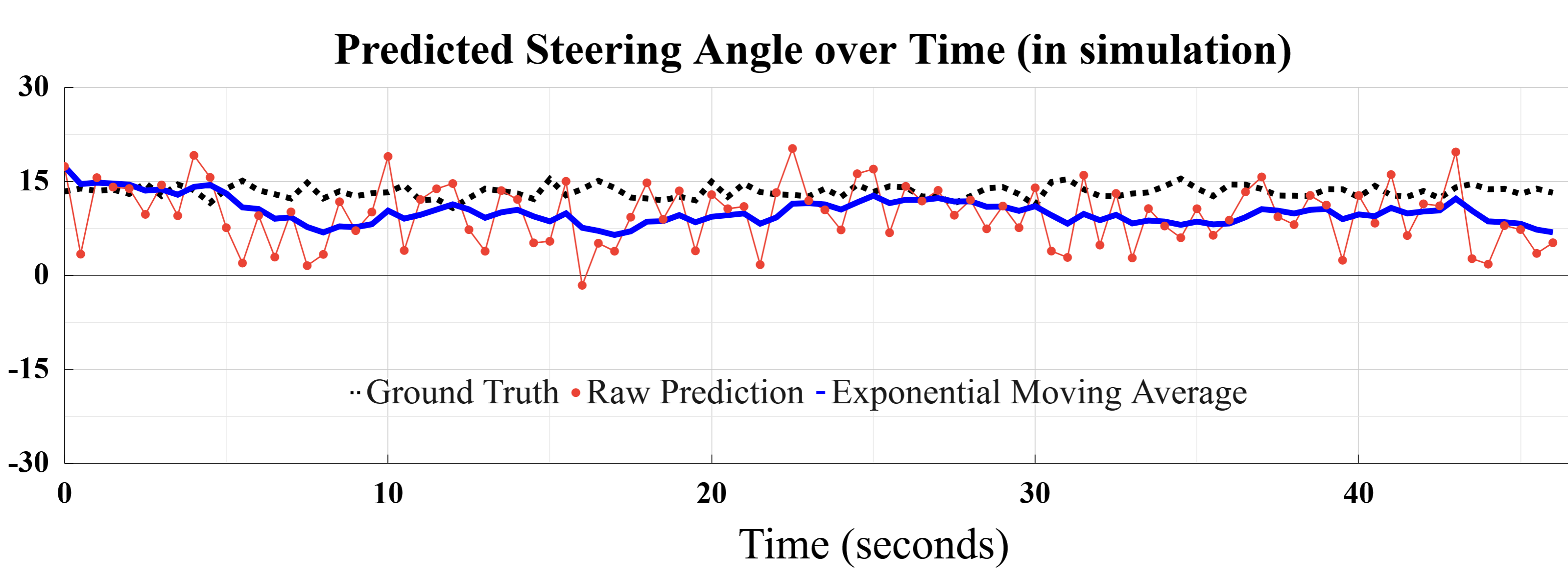}
    \caption{Model testing example using the GazelleSim environment. y-axis represents the steering angle in degrees.}
    \label{fig:field}
\end{figure}

The autonomous driving testing pipeline operates as follows:

\begin{enumerate}
    \item \textbf{Image Acquisition:} Simulated camera images are captured from the 2D environment, representing the vehicle's forward perspective.
    \item \textbf{Model Inference:} The captured image is fed into the trained model, which predicts a steering angle. The predicted steering angle represents the desired direction of the vehicle.
    \item \textbf{Steering Control:} The moving average of the predicted steering angle is translated into a control command that adjusts the vehicle's trajectory within the simulation. Then steering angle is converted into a control signal that influences the vehicle's direction and speed. For real-world testing, this control signal can be sent to the real vehicle's steering and throttle actuators instead.
    \item \textbf{Closed-Loop Feedback:} This pipeline loops back to image acquisition, model inference, and steering control, creating a closed-loop control system that is capable of driving the vehicle in real-time using onboard compute.
\end{enumerate}


\section{Experimental Results}

Table~\ref{tab:testarch} presents the units per layer of each of the four trained and optimized models. Layers that are dynamically optimized using ES with the 1/5 success rule are labeled in a bolded font, while non-bolded layers are static baseline PilotNET architecture layers modified according to Table~\ref{tab:aliases}.

\newcommand{\gr}{ \(\xrightarrow{}\) }

\begin{table}[ht]
    \centering
    \caption{Model Architectures after ES-Optimization}
    \begin{tabular}{c|c|c|c}

        \makecell{\textbf{Model}} & \makecell{\textbf{Architecture}} & \makecell{\textbf{N}} & \makecell{\textbf{M}}  \\
        \hline
        Baseline       & \makecell[l]{CNN: 24\gr36\gr48\gr64\gr64\gr\\
        DNN: 1164\gr100\gr50\gr10\gr1} & n/a & n/a \\
        \hline
        Optimized      & \makecell[l]{\textbf{CNN: 6\gr9\gr43\gr28\gr61\gr} \\
        \textbf{DNN: 2328\gr146\gr100\gr20}\gr1} & 2 & 4                   \\
        \hline
        Half-Size      & \makecell[l]{CNN: 12\gr18\gr24\gr32\gr32\gr        \\
        \textbf{DNN: 149\gr14\gr20}\gr1} & 4 & 6                             \\
        \hline
        Quarter-Size   & \makecell[l]{CNN: 6\gr9\gr12\gr16\gr16\gr          \\
        \textbf{DNN: 8\gr3}\gr1} & 3 & 6                                      \\
    \end{tabular}
    \label{tab:testarch}
\end{table}

These optimized models were tested in the GazelleSim environment to evaluate their real-time applicability. This test was conducted on a NVIDIA RTX A2000 Laptop GPU with 8GB VRAM and a lower power limit (35W) than the A100 GPU used in training. Even though our models do not saturate the VRAM, the power limit and memory bandwidth bottlenecks real-time inference latency, simulating a low power onboard compute environment.

The results are presented in Table~\ref{tab:testresults}. The speeds are not to-scale with real-world speeds due to simulator calibration; rather, the results can represent model performance as vehicle speeds increase while inference capacity is held constant.

\begin{table}[ht]
    \centering
    \caption{Test Results}
    \begin{tabular}{l|c|r}
        \textbf{Test Duration \& Speed} & \textbf{Model} & \textbf{Distance (m)} \\
        \hline
        \multirow{4}{*}{10 min @ 2 m/s} & Baseline       & 1200.026              \\
                                        & Optimized      & 1200.072              \\
                                        & Half-Sized     & 1200.206              \\
                                        & Quarter-Sized  & 1200.169              \\
        \hline
        \multirow{4}{*}{60 min @ 2 m/s} & Baseline       & 7200.026              \\
                                        & Optimized      & 7200.177              \\
                                        & Half-Sized     & 7200.053              \\
                                        & Quarter-Sized  & 7200.056              \\
        \hline
        \multirow{4}{*}{60 min @ 4 m/s} & Baseline       & 14400.408             \\
                                        & Optimized      & Failed                \\
                                        & Half-Sized     & Failed                \\
                                        & Quarter-Sized  & 14400.400
    \end{tabular}
    \label{tab:testresults}
\end{table}

Table~\ref{tab:compute} details the performance and size metrics of each of the four models. The number of parameters and MSE and MAE (evaluated on the test split of the dataset) were direct output from the Keras API. The VRAM usage of the inference process was measured using the nvidia-smi command during the trials. The inference times were averaged over 1000 predictions.

\begin{table}[ht!]
    \centering
    \caption{Compute Utilization}
    \begin{tabular}{c|c|c|c|c|c}
        \textbf{Model} & \textbf{Params} & \textbf{Memory} & \textbf{MSE} & \textbf{MAE} & \textbf{Inference} \\
        & & \textbf{(MB)} & \textbf{(deg)} & \textbf{(deg)} & \textbf{Time (ms)} \\ 
        \hline
        Baseline       & 245M                & 936                  & 1.01         & 0.63 & 4.8        \\
        Optimized      & 250M                & 1050                 & 0.49         & 0.41 & 4.5        \\
        Half-Size      & 61.4M               & 420                  & 1.33         & 0.78 & 4.2        \\
        Quarter-Size   & 4.97M               & 146                  & 1.88         & 0.88 & 4.1        \\
    \end{tabular}
    \label{tab:compute}
\end{table}

\section{Discussion}
Across 5-10 ES runs for each model size we observed the 1/5th rule reliably sort through hyperparameters to output a positively optimized end model for all runs. No instability or failed optimization runs occurred therefore, this proof-of-concept experiment establishes a simple statistical significance for using ES with 1/5 success rule to optimize convolutional and fully connected dense neural network layers for steering angle prediction in autonomous driving systems and hence meets our goals.

The PilotNET baseline achieved reasonable accuracy but required significant computational resources due to its large model size (245M params, 936 MB). The baseline model was able to predict steering angles with an average error of 1.01 degrees, and a maximum error of 0.63 degrees both within reasonable real-world driving accuracy. This model had no problems with real-time inference in simulations at 2 m/s and 4 m/s proving its suitability as a baseline for our study. This opens the way for architecture optimization to improve efficiency without sacrificing performance.

The ES-optimized PilotNET model achieved the lowest error (MSE: 0.49, MAE: 0.41 degrees) while the model size increased by 5M parameters. While this shows that ES-optimization works to improve model performance, it is important to note that the reduced models (half-size and quarter-size) showed trade-offs. They consumed significantly fewer resources and their prediction accuracy decreased, with the quarter-size model exhibiting the highest error (MSE: 1.88, MAE: 0.88 degrees). Even though a minor increase of error was observed in the quarter-size model, it required just 15\% of the baseline model's VRAM for inference while achieving equivalent performance in real-time simulations. In real-world low speed self-driving applications, a small difference in error (\textless1.0 degree) with a significant size reduction is a valid trade-off.

The 4m/s trials revealed a hard inference rate threshold, with Optimized (250M params) and Half-Sized (61.4M params) failing despite superior theoretical compute capacity. Given the close inference times in Table~\ref{tab:compute}, it is clear that the bottleneck is not the model size or memory availability. It may be the case that the end-to-end image to steering pipeline in the simulator is not fast enough to handle a simulated speed of 4m/s in real-time.

The Quarter-Sized model completed the 60-minute trial with only 0.008m deviation from Baseline (14400.400m vs. 14400.408m), demonstrating ES's ability to preserve functionality during radical downsizing. Minor performance variations at 2m/s (\textless0.18\% across models) further confirm architectural equivalence under non-saturating conditions.

This highlights the importance of balancing model performance and resource usage in real-world applications. Hence, our study shows that using ES with the 1/5 success rule to optimize models for better performance or size reduction does yield practical results that improve application efficiency. In the future, we plan to:
\begin{enumerate}
    \item Deploy the optimized CNN models on the LTU ACTor autonomous driving platform and compare their performance in real-world scenarios. Our very recent preliminary testing shows simulation equivalent real-world performance. The car was able to steer around without leaving the path while using all four models. Further testing and data will be collected later this year.
    \item Source low-power and low-compute embedded platforms such as NVIDIA Jetson Nano to evaluate these optimized models on. This will directly benchmark real-time performance (inferred frames per second and latency) and scalability of using ES based optimization for cost-effective systems in autonomous vehicle apps.
    \item Add small datasets by recording driving data on various roads across the LTU campus, including different surfaces, path geometries, and lighting conditions. Techniques like few shot learning and meta continual learning can be explored to further improve model performance.
    \item Add Recurrent Neural Network layers to train models based on the past steering inputs. This may improve model stability and smoothness.
\end{enumerate}

\bibliographystyle{IEEEtran} 
\bibliography{references} 

\end{document}